\documentclass[conference]{IEEEtran}
\IEEEoverridecommandlockouts

\usepackage{cite}
\usepackage{amsfonts}
\usepackage{amsmath,graphicx,booktabs,amssymb,siunitx,makecell}
\usepackage{algorithmic}
\usepackage{graphicx}
\usepackage{textcomp}
\usepackage{xcolor}
\def\BibTeX{{\rm B\kern-.05em{\sc i\kern-.025em b}\kern-.08em
    T\kern-.1667em\lower.7ex\hbox{E}\kern-.125emX}}
\begin{document}

\title{Large Language Models Can Understanding Depth from Monocular Images

}

\author{
\IEEEauthorblockN{1\textsuperscript{st} Zhongyi Xia}
\IEEEauthorblockA{\textit{College of Applied Technology} \\
\textit{Shenzhen University}\\
Shenzhen, China \\
2110413018@email.szu.edu.cn}
\and
\IEEEauthorblockN{2\textsuperscript{nd} Tianzhao Wu}
\IEEEauthorblockA{\textit{College of Applied Technology} \\
\textit{Shenzhen University}\\
Shenzhen, China \\
2110413016@email.szu.edu.cn}
\and
\IEEEauthorblockN{3\textsuperscript{rd} Given Name Surname}
\IEEEauthorblockA{\textit{dept. name of organization (of Aff.)} \\
\textit{name of organization (of Aff.)}\\
City, Country \\
email address or ORCID}
}

\maketitle

\begin{abstract}
Monocular depth estimation is a critical function in computer vision applications. This paper shows that large language models (LLMs) can effectively interpret depth with minimal supervision, using efficient resource utilization and a consistent neural network architecture. We introduce LLM-MDE, a multimodal framework that deciphers depth through language comprehension. Specifically, LLM-MDE employs two main strategies to enhance the pretrained LLM's capability for depth estimation: cross-modal reprogramming and an adaptive prompt estimation module. These strategies align vision representations with text prototypes and automatically generate prompts based on monocular images, respectively. Comprehensive experiments on real-world MDE datasets confirm the effectiveness and superiority of LLM-MDE, which excels in few-/zero-shot tasks while minimizing resource use. The source code is available.
\end{abstract}


\section{Introduction}


Monocular depth estimation (MDE) is essential for applications such as autonomous driving, where accurate environmental perception is critical for safety. Traditional MDE methods, based on manually designed features and geometric models, frequently underperform in complex scenarios. Recent advancements in deep learning (DL) have revolutionized MDE \cite{monodepth,AdaBinsDE,Lite-Mono}, offering robust performance without the constraints of physics or the need for resource-intensive feature engineering.

DL-based MDE techniques are divided into two categories based on learning strategies: supervised\cite{Xu2017MultiscaleCC,8546170,LIU2020107112} and unsupervised \cite{9803821,swinDepth} methods. Supervised methods require large labeled datasets and deliver impressive performance but are resource-intensive. In contrast, unsupervised methods use unlabeled data to facilitate effective knowledge transfer with minimal supervision. However, both strategies face three main challenges: (1) reliance on specialized neural architectures, requiring custom models for specific tasks, which reduces flexibility;  (2) the need for explicit information in certain scenarios, dependent on pre-trained pose estimation networks for scene-specific knowledge, limiting performance;  (3) dependency on precise data labeling, a premise rarely questioned in unsupervised methods despite minimal supervision.Therefore, developing a unified MDE framework that supports flexible performance with minimal supervision and independence from complex, tailor-made model architectures is crucial.

This paper demonstrates that pretrained large language models (LLMs) can effectively understand depth from monocular images. We introduce the \textbf{L}arge \textbf{L}anguage \textbf{M}odel for \textbf{M}onocular \textbf{D}epth \textbf{E}stimation (dubbed LLM-MDE), a multi-modal framework that interprets depth via language understanding. LLM-MDE integrates two primary strategies to improve depth perception: cross-modal reprogramming and an adaptive depth prompt generation. The former aligns visual representations from monocular images with text prototypes from a comprehensive vocabulary library, enhancing feature extraction for LLM input. The latter strategy generates and tokenizes prompts from monocular images for LLM processing. These approaches significantly improve LLM insights into monocular depth estimation. Our contributions are four-fold:
\begin{itemize}
    \item This study represents the first exploration of pre-trained large language models (LLMs) for monocular depth estimation. Empirical evidence demonstrates that LLMs can deliver depth information with minimal supervision.
    \item We introduce LLM-MDE, a unified multimodal framework utilizing LLMs for monocular depth estimation. It integrates cross-modal reprogramming and an adaptive depth prompt generation module to enhance LLM insights into depth with minimal supervision and resource.
    \item We introduce cross-modal reprogramming and adaptive depth estimation. The former aligns monocular image and text prototypes, while the latter automatically generates depth prompts to enhance estimation insights.
    \item Extensive experiments on the real-world MDE dataset demonstrate the effectiveness and superiority of our LLM-MDE, which performs well on few-/zero-shot tasks.
\end{itemize}
We highlight that LLM-MDE is not for competitive purposes but rather serves as an exploratory tool for depth estimation, especially in scenarios with limited supervision/resources or where complex neural architectures are not required.

\section{Methodology}
The structure of our LLM-MDE is illustrated in Fig.~\ref{fig:framework}. It combines two pretrained models: a Vision Transformer (ViT) that extracts visual representations from images and an LLM that performs depth estimations. We introduce two strategies: cross-modal reprogramming and adaptive depth prompt generation, which enhance the LLM's depth estimation capabilities. Features from these strategies are fused into the LLM via an adaptive head for accurate depth estimation. Further details will be provided subsequently.



\subsection{Cross-modal Reprogramming between Vision and Text}
LLM pretrained on extensive natural language datasets demonstrate superior sequence modeling and generalization capabilities. However, differences between text and image data prevent direct application of LLMs to image representation tasks. Monocular images also cannot be directly edited or described losslessly in natural language, posing significant challenges for using LLMs to understand them without intensive fine-tuning. To address this, we introduce a cross-modal reprogramming strategy that combines visual representations of monocular images with latent semantic information from large-scale textual corpora, enhancing the LLM's ability to perceive, understand, and interpret vision representations. Specifically, we used pre-trained word embedding $\mathbf{E} \in \mathbb{R}^{V \times D}$ in the LLM backbone, where $V$ and $D$ denote the vocabulary size and dimension. Nevertheless, there is no prior knowledge indicating which text tokens are directly relevant with monocular image representation. Thus, we maintain a small collection of text prototypes by linearly transformation $\mathbf{E}$, denoted as $\mathbf{E}' \in \mathbb{R}^{V' \times D}$, where $V' << V$. Text prototypes learn connecting to represent the local patch information (e.g., ``extremely close`` for vision representation) without leaving the space where the language model is pre-trained. We achieve the proposed Cross-modal Reprogramming via a multi-head attention layer. For each haed $k = \left\{1, \cdots, K \right\}$, we define query matrices $\mathbf{Q}^{(i)}_k = \hat{\mathbf{X}}^{(i)}_P W^Q_k$, key matrices $\mathbf{K}^{(i)}_k =\mathbf{E}' W^K_k$, and value matrices $\mathbf{V}^{(i)}_k = \mathbf{E}' W^V_k$, where $W^Q_k \in \mathbb{R}^{d_m \times d}$ and  $W^K_k, W^K_V \in \mathbb{R}^{D \times d}$. Specifically, $D$ is the hidden dimension of the pretrained LLM, and $d = \frac{d_m}{K}$. Then, the cross-modal reprogramming can be formulated as:
\begin{equation}
\begin{aligned}
    \mathbf{F}^{(i)}_k &= \texttt{Reprogramming}(\mathbf{Q}^{(i)}_k, \mathbf{K}^{(i)}_k, \mathbf{V}^{(i)}_k) \\
    &= \textsc{Softmax}(\frac{\mathbf{Q}^{(i)}_k \mathbf{K}^{(i)}_k}{\sqrt{d_k}})\mathbf{V}^{(i)}_k
\end{aligned}
\end{equation}
Finally, by aggregating the features $\mathbf{F}^{(i)}_k \in \mathbb{R}^{D' \times d}$ from each head, we obtain $\mathbf{F}^{(i)} \in \mathbb{R}^{D' \times d_m}$, where $D'$ is the output dimension of Cross-domain Reprogramming. These are then linearly projected to fuse with the representation from the prompt representation detailed below.


\subsection{Adaptive Depth Prompts Generation Module}
To strength the insight of depth understanding of pretrained LLMs without additional structures or internal modifications, we introduce the Adaptive Depth Prompt Generation Module (APG). The APG autonomously generates statistical prompts for monocular images, improving depth comprehension. This module integrates prompt generation and representation, producing prompts from four perspectives: Dataset, Task, Pixel, and Class. The Dataset and Task components generate concise dataset information and task descriptions. The Pixel component creates prompts using pixel-level statistics like minimum, maximum, and median values from the monocular image. Class assigns a unique label to each image based on pixel value distribution across seven categories: ``giant``, ``extremely close``, ``close``, ``not in distance``, ``a little remote``, ``far``, and ``unseen``. The generated prompts are then processed by a pretrained tokenizer to yield textual representation.

\subsection{Depth Projection from Adaption Head}
To transform language representations into depth information, we introduce the Adaptation Head based on the ResNet architecture for feature refinement and depth projection. The Adaptation Head employs the UpsampleBN module, integrating convolution, batch normalization, and Leaky ReLU with residual connections. The process starts by adjusting input features with a linear layer, followed by three UpsampleBN operations to enhance spatial resolution and feature representation. This expands feature maps to capture fine details and increase the receptive field. A final \texttt{Sigmoid} normalizes the output, producing the depth map.

\subsection{Lightweight Operations and Optimization}
Tuning pre-trained ViTs and LLMs for visual representation and depth estimation remains resource-intensive, posing significant challenges in low-resource settings. To address this, we introduce lightweight operations throughout the framework to balance cost and performance. Specifically, we adopt low-rank adaptation (LoRA)~\cite{hu2021lora} for each attention block within the ViT and LLM, which efficiently updates parameters by modifying only a small subset of weights, preserving the original model structure and knowledge. The implementation of LoRA involves using the original weight matrix $W \in \mathbb{R}^{d \times d}$ and adding the product of lower-order matrices as:
\begin{equation}
    W' = W + A\times B, \quad \textit{where}\quad A \in \mathbb{R}^{d \times r}, B \in \mathbb{R}^{r \times d},
\end{equation}
where $r$ denotes the rank value, and $A$ and $B$ are low-rank matrices with dimensions smaller than $W$ ($r \ll d$), ensuring a low parameter count in the tuning process. For optimization, we used the scale-invariant squared loss (SSI) for monocular depth estimation is formulated as:
\begin{equation}
    L(\theta) = \frac{1}{n} \sum_{i=1}^n \left( \log d_i - \log \hat{d}_i - \frac{1}{n} \sum_{j=1}^n (\log d_j - \log \hat{d}_j) \right)^2
\end{equation}
where $\theta$ represents the model unfrozen parameters, $d_i$ is the true depth value for the $i$-th sample, $\hat{d}_i$ is the predicted depth value for the $i$-th sample, and $n$ is the number of samples.

\section{Experiments}
We conducted evaluation on Ubuntu 22.04 server, equipped with an Intel Xeon Silver 4210R CPU and an NVIDIA GeForce RTX 3090Ti GPU (24 GB RAM). Key hyperparameters were set as follows: a patch size of 16, training resolution of 224, a dropout rate of 0.1, a batch size of 16, and the \texttt{AdamW} optimizer with an initial learning rate $1e^{-5}$. We utilized the NYU raw dataset, which comprises images with a resolution of $640 \times 480$, in all experiments due to its generalizability. We used the ViT-base and 12-layer BERT throughout all experiments. During training, we conducted 50 epochs with an early-stopping strategy that halts training if the validation loss does not decrease for 5 consecutive rounds. Additionally, we applied a cosine annealing strategy to the learning rate to prevent overfitting. We closely adhere to the experimental protocol outlined by Ranftl et al.\cite{ranftlDPT} Specifically, we utilize Root Mean Squared Error (RMSE), Absolute Relative Error (Abs Rel), Squared Relative Error (Sq Rel), Logarithmic Root Mean Squared Error (Log RMSE), and accuracy as our evaluation metrics.

\subsection{Few-Shot and Zero-Shot Experiments}

\begin{figure}[tbh]
  \centering
  \includegraphics[width=.485\textwidth]{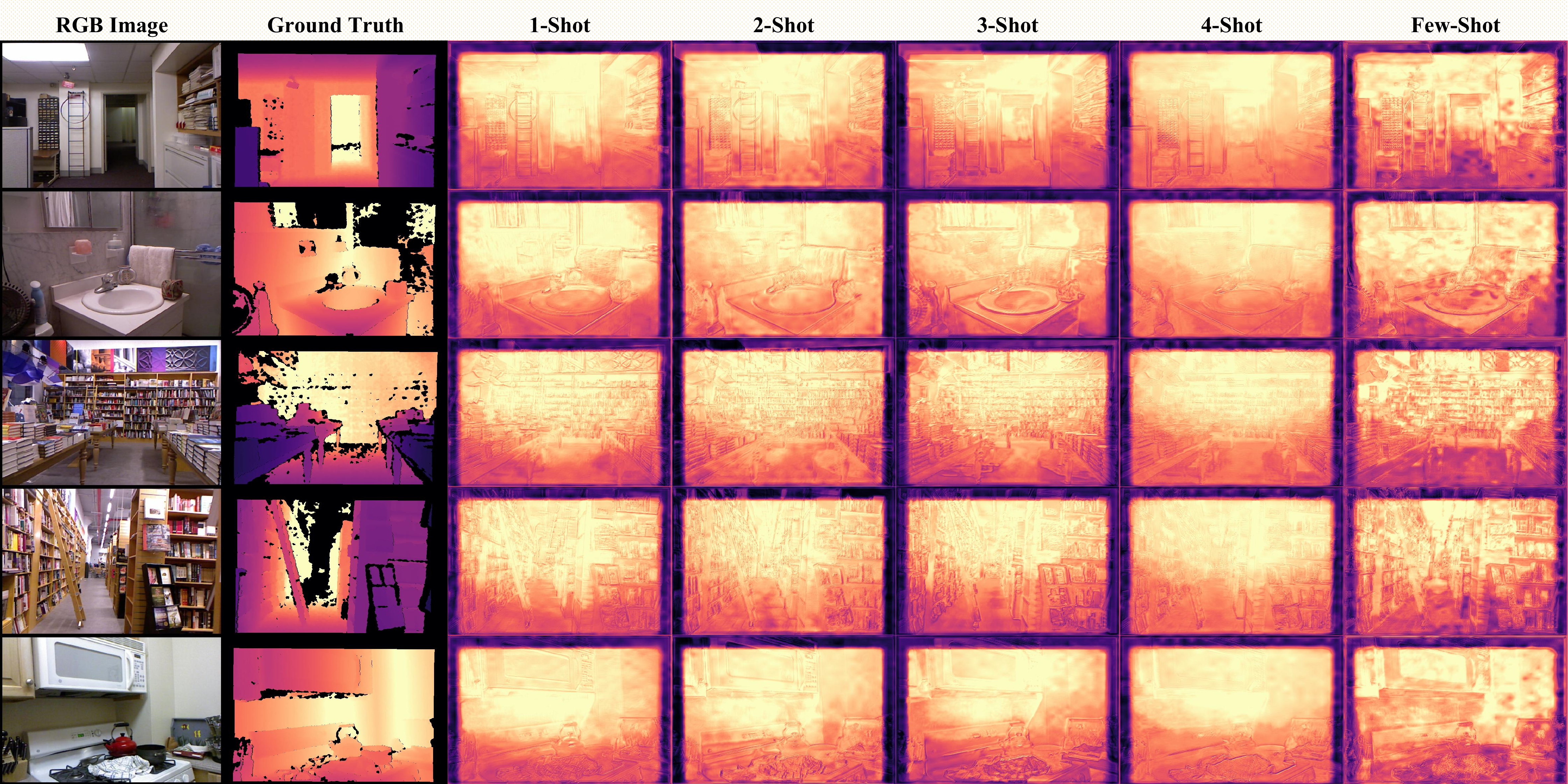}
\caption{Visual results of the few-shot experiments with limited resources.}
\label{fig:few-shot-experiment-all}
\end{figure}

To demonstrate the effectiveness of LLM-MDE in resource-limited settings, we executed Few-shot and Zero-shot experiments. The results, as depicted in Tab.\ref{tab:few-shot-experiment-all} and Fig.\ref{fig:few-shot-experiment-all}, show that the Few-Shot experiments were divided into five groups. The initial four groups ranged from 1-Shot to 4-Shot, with each group containing 50 to 100 images. The fifth group, labeled as Few-Shot, comprised a single randomly selected image from each scene type, totaling 28 images. Incremental increases in the number of shots led to substantial reductions in various losses and enhancements in detail resolution, exemplified by improved texture depiction in bookshelves in the third and fourth images, and more accurate delineation of invalid areas in the second and fourth images.

\begin{table}[tbh]
\centering
\caption{Few-shot experiment results with limited resources. \textbf{Bold} denotes the best.}
\label{tab:few-shot-experiment-all}
\begin{tabular}{@{}lccccc@{}}
\toprule
\textbf{Class Labels} & \textbf{1-Shot} & \textbf{2-Shot} & \textbf{3-Shot} & \textbf{4-Shot} & \textbf{Few-Shot} \\
\midrule
Bedroom & $\checkmark$ & $\checkmark$ & $\checkmark$ & $\checkmark$ & $\checkmark$ \\
Bathroom & & $\checkmark$ & $\checkmark$ & $\checkmark$ & $\checkmark$ \\
Diningroom & & & $\checkmark$ & $\checkmark$ & $\checkmark$ \\
Kitchen & &  &  &$\checkmark$ &$\checkmark$ \\
Remaining classes & & & & & $\checkmark$\\
\midrule
\textbf{Conclusion} & & & & &  \\
\midrule
RMSE & 0.285 & 0.267 & 0.259 & \textbf{0.242} & 0.253 \\
Abs Rel & 0.741 & 0.707 & 0.669 & \textbf{0.627}  & 0.639 \\
Sq Rel & 0.318 & 0.289 & 0.265 & \textbf{0.234}  & 0.247 \\
Log RMSE & 0.542 & 0.526 & 0.508 & \textbf{0.488}  & 0.498 \\
$\delta_1$ & 0.365 & 0.389 & 0.394 & \textbf{0.415}  & 0.402 \\
$\delta_2$ & 0.574 & 0.591 & 0.612 & \textbf{0.637}  & 0.625 \\
$\delta_3$ & 0.731 & 0.745 & 0.765 & \textbf{0.783}  & 0.777 \\
\bottomrule
\end{tabular}
\end{table}

\begin{figure}[tbh]
  \centering
  \includegraphics[width=.485\textwidth]{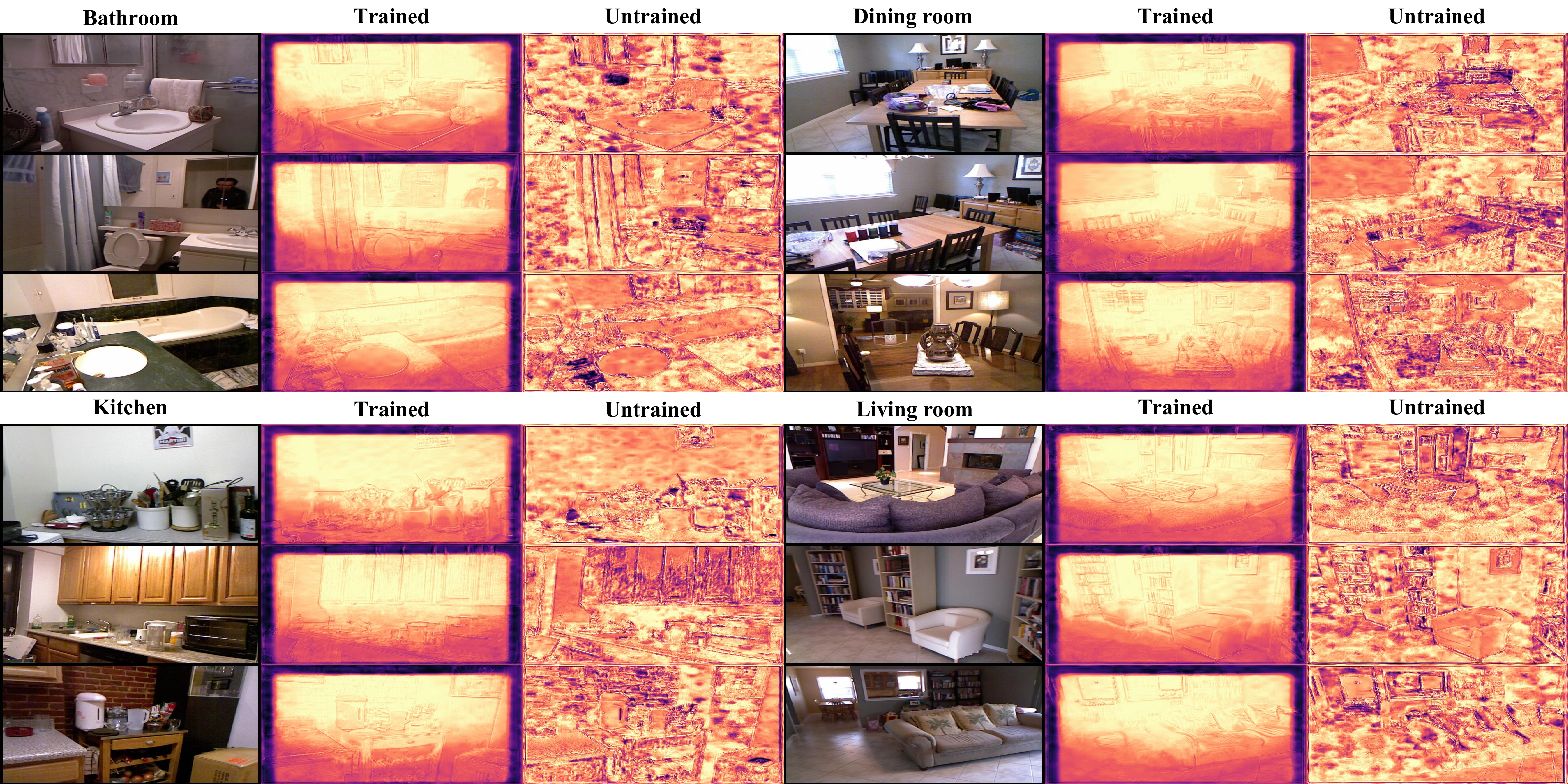}
    \caption{Visual results of the cross-domain zero-shot experiments.}
    \label{fig:zero-shot}
\end{figure}

As shown in Tab.~\ref{tab:cross-domain-zero-shot-experiments} and Fig.~\ref{fig:zero-shot}, Zero-shot experiments trained on one scene and tested across four unseen types demonstrate LLM-MDE's generalization. Although untrained on these scenes, the model achieved low loss values, highlighting its robustness. Fig.~\ref{fig:zero-shot} shows that without training, the model captures only partial texture details and inaccurately estimates depth. After cross-domain training, visual results improve significantly. For instance, in the Living Room scene, the trained model accurately identifies the depth of the sofa, floor, and distant objects, while in the Bathroom scene, it captures the texture and depth of the sink and bathtub effectively.

\begin{table}[tbh]
  \centering
  \caption{Cross-domain zero-shot experiments results. \textbf{Bold} denotes the best.}
  \label{tab:cross-domain-zero-shot-experiments}
  
  \sisetup{
    table-number-alignment = center,
    detect-weight,
    mode=text
  }
  
  \setlength{\tabcolsep}{4pt} 
  \renewcommand{\arraystretch}{1.2} 
  
  \begin{tabular}{
    >{\bfseries\raggedright}p{2cm} 
    S[table-format=1.3]
    S[table-format=1.3]
    S[table-format=1.3]
    S[table-format=1.3]
  }
    \toprule
    Type &
    \textbf{RMSE} &
    \textbf{Abs Rel} &
    \textbf{Sq Rel} &
    \textbf{Log RMSE} \\
    \midrule
    Bathroom &\textbf{0.287} & \textbf{0.724} & \textbf{0.319} & \textbf{0.529} \\
    Dining room & 0.338 & 1.022 & 0.467 & 0.688 \\
    Kitchen & 0.345 & 1.100 & 0.537 & 0.699 \\
    Living room & 0.310 & 0.835 & 0.348 & 0.604 \\
    \bottomrule
  \end{tabular}
\end{table}

\subsection{Ablation Experiments}

\begin{figure}[tbh]
  \centering
  \includegraphics[width=.485\textwidth]{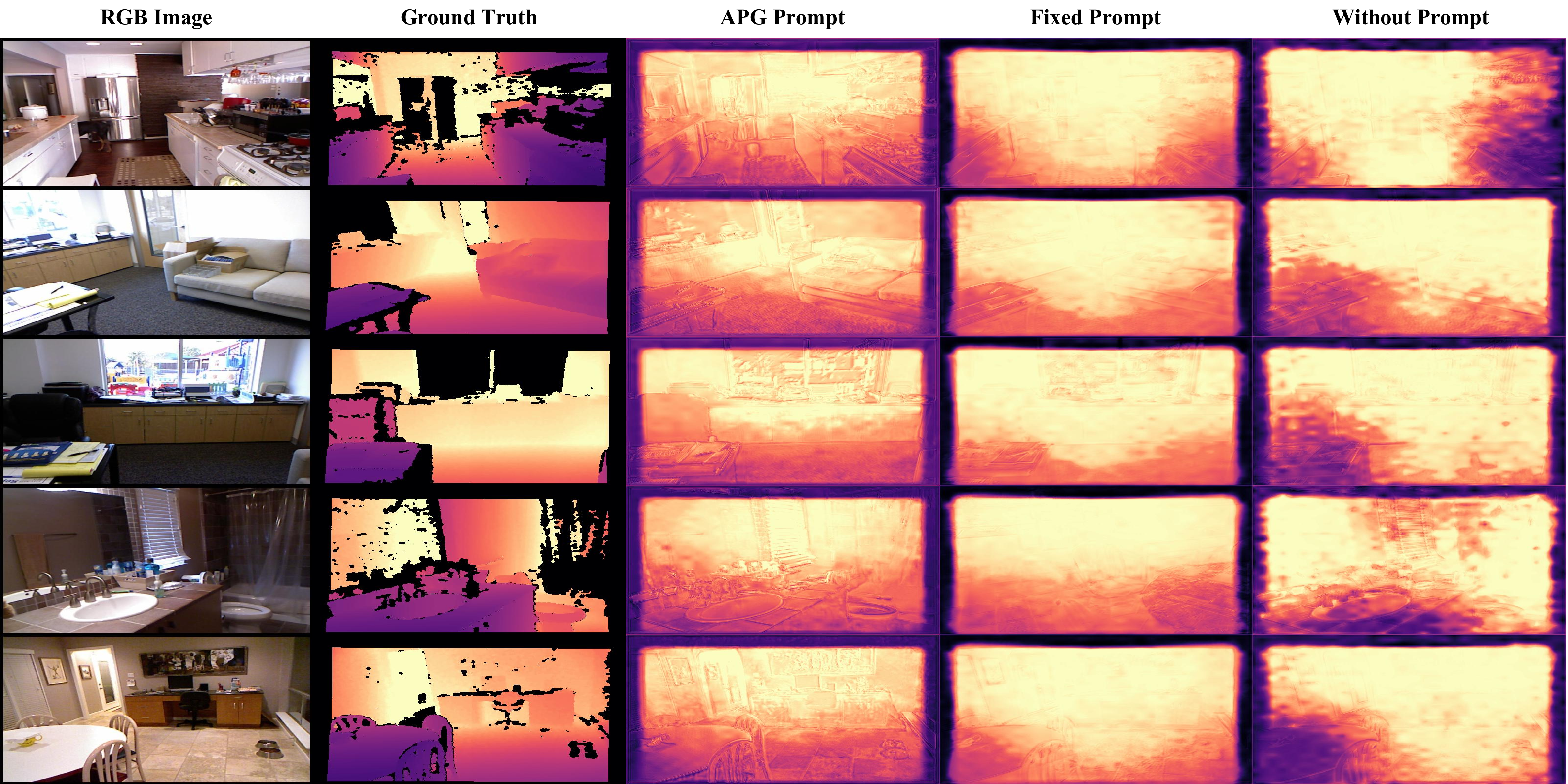}
    \caption{Visual results of the prompts ablation study.}
\label{fig:prompts}
\end{figure}

To demonstrate the effectiveness of APG and Fixed Prompts in depth estimation, we conducted an ablation study, the results of which are shown in Tab.~\ref{tab:prompt} and Fig.~\ref{fig:prompts}. The model without prompts exhibited the highest loss, marked by significant noise and artifacts. Conversely, Fixed Prompts significantly reduced loss, lowering RMSE and Abs Rel by 31.4\% and 43.4\%, respectively, and reducing artifacts. APG Prompts showed superior performance, minimizing artifacts and enhancing textural details. For instance, in Fig.~\ref{fig:prompts}, the APG Prompt effectively captures the texture of the sink in the third column, fourth row, and the details of the table and chairs in the third column, fifth row. We also conducted qualitative and quantitative analyses to confirm these results, verifying the superior efficacy of APG Prompts in improving depth estimation accuracy.

\begin{table}[tbh]
\centering
\caption{Ablation results on prompting mechanism. LLM-MDE-A: APG Prompts only. LLM-MDE-B: Fixed Prompts only. LLM-MDE-C: Without Prompts. \textbf{Bold} denotes the best.}
\label{tab:prompt}
\begin{tabular}{ccccc}
\toprule
\textbf{Prompts} & \textbf{RMSE} & \textbf{Abs Rel} & \textbf{Sq Rel} & \textbf{Log RMSE}\\
\midrule 
\textbf{LLM-MDE-A}& \textbf{0.206} & \textbf{0.448} & \textbf{0.125} & \textbf{0.426}\\
\textbf{LLM-MDE-B} & 0.214 & 0.461 & 0.132 & 0.441\\
\textbf{LLM-MDE-C} & 0.312 & 0.814 & 0.363 & 0.579\\
\bottomrule 
\end{tabular}
\end{table}

\begin{figure}[tbh]
  \centering
  \includegraphics[width=.485\textwidth]{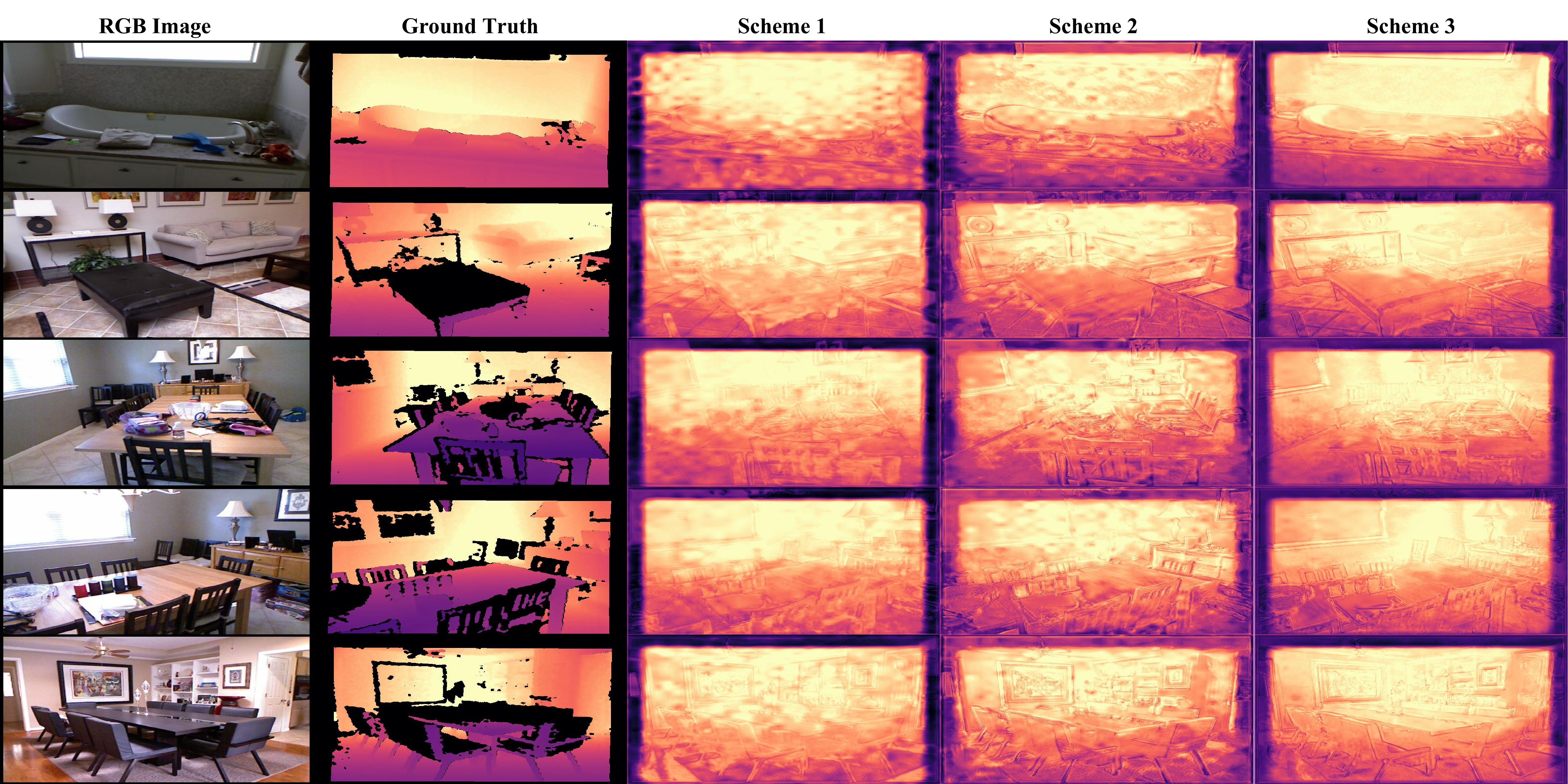}
    \caption{Visual results of the LoRA fine-tuning experiments.}
\label{fig:LoRA Ablation}
\end{figure}

As shown in Tab.~\ref{tab:LoRA Ablation} and Fig.~\ref{fig:LoRA Ablation}, we conducted an ablation study to validate the effectiveness of the LoRA fine-tuning strategy for depth estimation. Scheme 1, which uses Frozen ViT and Frozen LLM as a control group, exhibited high model losses and significant artifacts. Scheme 2, replacing Frozen ViT with LoRA ViT, reduced artifacts and decreased Abs Rel and Sq Rel by 30.0\% and 47.0\%, respectively. Scheme 3, further substituting Frozen LLM with LoRA LLM, achieved the lowest losses, with Abs Rel and Sq Rel decreasing by 40.0\% and 61.0\%, respectively, effectively eliminating artifacts and providing more accurate predictions.

\begin{table}[tbh]
\centering
\caption{Ablation results on LoRA fine-tuning experiments.. Scheme 1: Frozen ViT and Frozen LLM. Scheme 2: LoRA ViT and Frozen LLM. Scheme 3: LoRA ViT and LoRA LLM. \textbf{Bold} denotes the best.}
\label{tab:LoRA Ablation}
\begin{tabular}{ccccc}
\toprule 
\textbf{Components} & \textbf{RMSE} & \textbf{Abs Rel} & \textbf{Sq Rel} & \textbf{Log RMSE} \\
\midrule 
\textbf{Scheme 1}  & 0.288 & 0.748 & 0.320 & 0.549 \\
\textbf{Scheme 2}   & 0.218 & 0.522 & 0.171 & 0.449 \\
\textbf{Scheme 3} & \textbf{0.206} & \textbf{0.448} & \textbf{0.125} & \textbf{0.426} \\
\bottomrule 
\end{tabular}
\end{table}

\subsection{Hyper-parameter Sensitivity}
\begin{figure}[tbh]
  \centering
  \includegraphics[width=.485\textwidth]{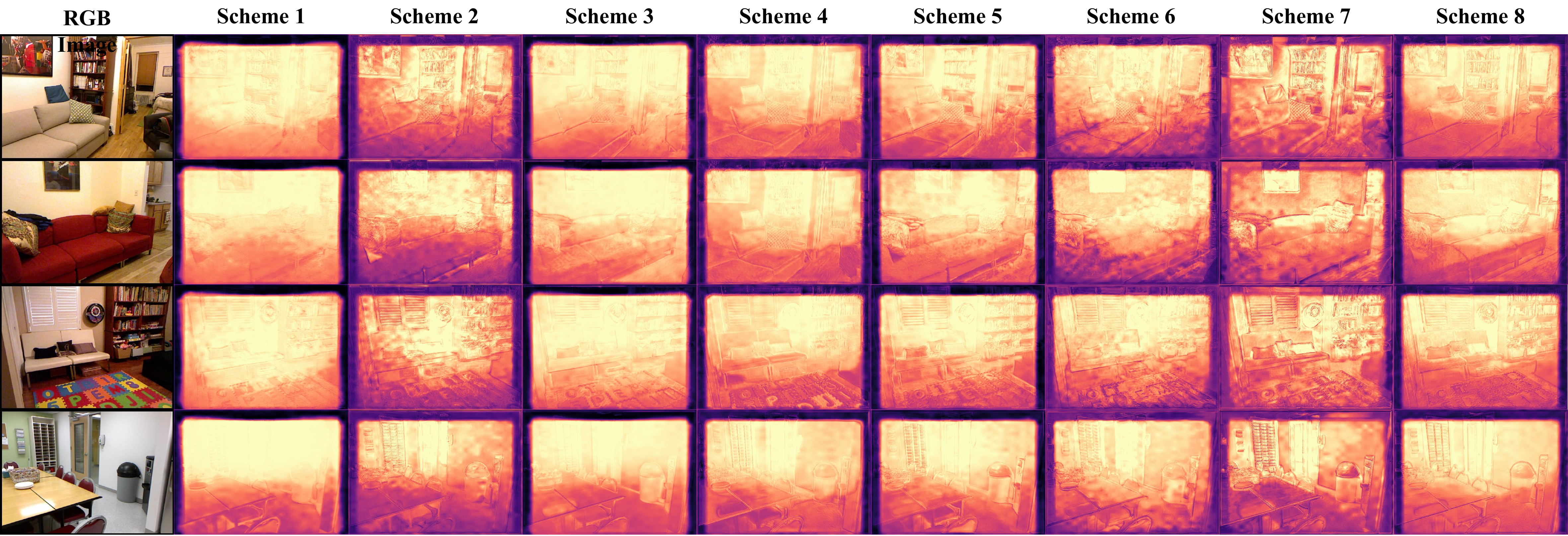}
    \caption{Visual results of the hyperparameter sensitivity fine-tuning experiments. The detailed information about 8 scheme can be found at Tab.~\ref{fig:Hyper-parameter Sensitivity}.}
\label{fig:Hyper-parameter Sensitivity}
\end{figure}

Tab.~\ref{tab:Hyper-parameter Sensitivity} and Fig.~\ref{fig:Hyper-parameter Sensitivity} present the results of the LLM-MDE hyper-parameter sensitivity experiment involving various LoRA fine-tuning strategies. We used a controlled variable approach, adjusting the Alpha and Rank parameters of LoRA ViT and LoRA LLM, as well as batch size and learning rate, to study their impact on model accuracy. Schemes 1, 3, and 7 show that low Alpha and Rank values reduce LoRA's effectiveness: Scheme 1 shows less detailed predictions, while Scheme 7 has more artifacts. Schemes 3 and 6 demonstrate that very high Alpha and Rank values cause overfitting and poor generalization, leading to significant artifacts. Schemes 2 and 3 reveal that too much parameter adjustment freedom undermines training stability and increases losses and artifacts. Schemes 3, 5, and 8 indicate that smaller batch sizes reduce training stability and prediction accuracy, and increase losses. However, as Scheme 8 shows, very large batch sizes on small datasets can also impair accuracy.

\begin{table*}[tbh]
\centering
\caption{Results of the hyperparameter sensitivity fine-tuning experiments. \textbf{Bold} denotes the best.}
\begin{tabular}{lcccccccc}
\toprule
\textbf{Variable Name} & \textbf{Scheme 1} & \textbf{Scheme 2} & \textbf{Scheme 3} & \textbf{Scheme 4} & \textbf{Scheme 5} & \textbf{Scheme 6} & \textbf{Scheme 7} & \textbf{Scheme 8} \\
\midrule
Alpha (ViT) & 120 & 192 & 192 & 192 & 192 & 320 & 192 & 192 \\
Rank (ViT) & 60 & 192 & 96 & 96 & 96 & 160 & 96 & 96 \\
Rank (LLM) & 32 & 32 & 32 & 32 & 32 & 32 & 16 & 32 \\
Batch size & 32 & 32 & 32 & 32 & 16 & 32 & 32 & 48 \\
Learning rate & 2e-5 & 2e-5 & 2e-5 & 1e-4 & 2e-5 & 2e-5 & 2e-5 & 2e-5 \\
\midrule
\textbf{Conclusion} &  &  &  &  &  &  &  &  \\
\midrule
RMSE & 0.338 & 0.218 & \textbf{0.206} & 0.284 & 0.252 & 0.258 & 0.261 & 0.261 \\
Abs Rel & 0.880 & 0.496 & \textbf{0.448} & 0.743 & 0.593 & 0.632 & 0.657 & 0.669 \\
Sq Rel & 0.415 & 0.158 & \textbf{0.125} & 0.317 & 0.215 & 0.247 & 0.259 & 0.265 \\
Log RMSE & 0.607 & 0.440 & \textbf{0.426} & 0.541 & 0.518 & 0.499 & 0.509 & 0.507 \\
$\delta_1$ & 0.281 & \textbf{0.426} & 0.393 & 0.359 & 0.370 & 0.390 & 0.382 & 0.387 \\
$\delta_2$ & 0.494 & 0.678 & \textbf{0.708} & 0.574 & 0.631 & 0.623 & 0.611 & 0.610 \\
$\delta_3$ & 0.668 & 0.831 & \textbf{0.865} & 0.734 & 0.788 & 0.779 & 0.768 & 0.765 \\
\bottomrule
\end{tabular}
\label{tab:Hyper-parameter Sensitivity}
\end{table*}

\section{Conclusions}
In conclusion, this paper introduces LLM-MDE, a multi-modal framework that interprets depth through language understanding. LLM-MDE employs two main strategies to enhance depth perception: cross-modal reprogramming and an adaptive depth estimation module. The former aligns visual representations from monocular images with text prototypes from a comprehensive vocabulary, improving feature extraction for LLM input. The latter generates and tokenizes prompts from images for LLM processing. These methods significantly enhance monocular depth estimation insights. Extensive experiments on the real-world MDE dataset demonstrate the effectiveness and superiority of our LLM-MDE.

\clearpage
\bibliographystyle{IEEEbib}
\bibliography{IEEE-conference-template}
\end{document}